\documentclass[a4paper, 10pt, conference]{ieeeconf}      
\usepackage{FG2020}

\FGfinalcopy 

\usepackage{times}
\usepackage{epsfig}
\usepackage{graphicx}
\usepackage{amsmath}
\usepackage{amssymb}

\usepackage{cite}
\usepackage{amsmath,amssymb,amsfonts}
\usepackage{algorithmic}
\usepackage{graphicx}
\usepackage{textcomp}
\usepackage{xcolor}
\usepackage{multirow}

\IEEEoverridecommandlockouts                              
\newcommand{\comment}[1]{}
\newcommand{\etal}{\textit{et al.}}

\def\FGPaperID{11} 

\title{\LARGE \bf
FineHand: Learning Hand Shapes for American Sign Language Recognition
}

\author{\parbox{16cm}{\centering
		{Al Amin Hosain, Panneer Selvam Santhalingam, Parth Pathak, Huzefa Rangwala and Jana Ko\v{s}eck\'a}\\
		{\normalsize
			Department of Computer Science, George Mason University, Fairfax, Virginia, USA}}
	\thanks{This work was supported by Google Faculty Research Award}
}

\begin{document}

\ifFGfinal
\thispagestyle{empty}
\pagestyle{empty}
\else
\author{Anonymous FG2020 submission\\ Paper ID \FGPaperID \\}
\pagestyle{plain}
\fi
\maketitle

\begin{abstract}
American Sign Language recognition is a difficult gesture recognition problem, characterized by fast, highly articulate gestures.
These are comprised of arm movements with different hand shapes, facial expression and head movements. 
Among these components, hand shape is the vital, often the most discriminative part of a gesture. 
In this work, we present an approach for effective learning of hand shape embeddings, which are discriminative for ASL gestures. 
For hand shape recognition our method uses a mix of manually labelled hand shapes and high confidence predictions to train deep convolutional neural network (CNN). The sequential gesture component is captured by recursive neural network (RNN) trained on the embeddings learned in the first stage. We will demonstrate that higher quality hand shape models can significantly improve the accuracy of final video gesture classification in challenging conditions with variety of speakers, different illumination and significant motion blurr.  We compare our model to alternative approaches exploiting different modalities and representations of the data and show improved video gesture recognition accuracy on GMU-ASL51 benchmark dataset. 
\end{abstract}

\section{Introduction}

Despite numerous efforts in the past addressing different components of American Sign Language (ASL) gesture recognition, automated sign language parsing in the wild remains challenging. The presented work is motivated by the need to design interfaces that can enable interactions between a Deaf and Hard-of-Hearing (DHH) user and a digital assistant (e.g. Amazon Echo, Google Now). Intelligent virtual assistant devices are becoming ubiquitous and the types of services they offer continues to expand. They can help users in answering questions, managing schedules, describing weather and many more. However, most of these devices are voice controlled. Hence, DHH people are deprived of the benefits from using these assistants. 

An ASL sign is performed by a combination of hand gestures, facial expressions and postures of the body. Further, the sequential motion of specific body locations (such as hand-tip, neck and arm) provide informative cues about a sign. 



\begin{figure}[h]
	\centering
	\includegraphics[clip, trim= 0.0cm 19.0cm 5.5cm 0.5cm,width=\linewidth]{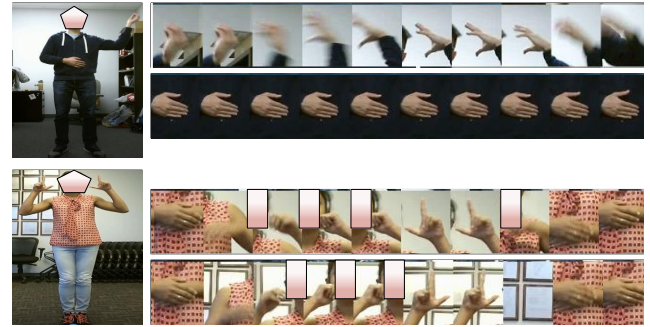}
	\caption{Subjects performing different ASL signs and corresponding hand-shape pattern. Faces are masked for privacy concern (Note: All faces are masked in this paper for the same reason).}
	\label{fig:sign_close}
\end{figure}
 
In this work we consider the problem of classification of individual ASL signs captured by short video snippets in unrestricted settings, by multiple signers. 
We focus on learning effective hand shape representations robust to changes of style, motion blur and illumination (see Figure~\ref{fig:sign_close}). To localize the hands, we exploit recent advances in human pose estimation methods from video, which are effective in estimating 2D hand joint locations ~\cite{cao2018openpose,wei2016cpm,xiu2018poseflow}. 
In order train discriminative hand shape model, frame level annotation 
of hand shapes classes is required. Such labelling tedious and time consuming process. 
For this reason, we first manually select a small set of training hand shape examples from each video gesture to train the initial model  and keep increasing this collection using the predictive power of deep convolutional neural networks (CNNs).  We will show that, this type of supervision can benefit gesture recognition accuracy while keeping the manual annotation  effort at minimum. 
By training a network for hand shape classification, we obtain   discriminative hand shape embeddings as penultimate layer of the network.  These are subsequently used to learn sequential dynamics of video gestures video using recursive neural network (RNN)~\cite{DBLP:journals/corr/Lipton15}.
In summary, the contributions of this work are:


\begin{itemize}
	\item We propose an iterative learning mechanism to train a deep CNN to learn robust hand shape embeddings.
	\item We implement sequential RNN models for video gesture recognition using the learned hand shape embeddings.
	\item We show evaluation of our method varying different factors such as fraction of hand-shape supervision and single hand vs both hands and compare it with other multi-modal methods for ASL gesture  recognition.
	\item We create and release per frame hand-shape annotations for GMU-ASL51 dataset. This will pave the way for rigorous sequential machine learning on the dataset.
\end{itemize}
We will demonstrate superior performance of the proposed model
on the GMU-ASL51 dataset~\cite{gmuASL} and compare it quantitatively with several baseline approaches which use different representations and different sensing modalities. 


\section{Related Work}
Previous work on sign language recognition focused either on video gestures, finger spelling or sentence parsing and deployed variety of sensing modalities. With the exception of few many benchmarks were  acquired in controlled laboratory settings, with small variation in speakers, clothing and lighting or used speaker augmentation to 
make the hand detection, tracking and recognition more robust. 

In~\cite{copycat_zafrulla} authors developed  HMM based framework for ASL phrase verification and subjects used colored gloves during data collection. The follow up work used 
Kinect skeletal data and accelerometers worn on hands~\cite{Zafrulla:2011:ASL:2070481.2070532}. 
Starner\cite{Starner:1998:RAS:297806.297828} demonstrated two HMM based approaches to recognize sentence level ASL using a single camera to track the user's unadorned hands, but used carefully chosen clothing and backgrounds for hand detection and tracking purposes. 

More recently, 
approaches based on deep neural network have attracted much attention in modeling sign language and enabled to bypass the cumbersome feature engineering stage.
With the advent of deep learning techniques, several approaches for ASL gesture recognition have been developed for learning discriminative spatio-temporal features from video~\cite{7177428,Ji:2013:CNN:2412386.2412939,Sun:2015:LSV:2753829.2629481} and applied to finger spelling recognition. 
Huang~\cite{7177428} showed the effectiveness of using Convolutional neural network (CNN) with RGB video data for sign language recognition. Three dimensional CNN has been used to extract spatio-temporal features from the video in~\cite{Ji:2013:CNN:2412386.2412939}. Similar architecture was implemented for Italian gestures \cite{978-3-319-16178-5_40}. 
Sun \etal \cite{Sun:2015:LSV:2753829.2629481}
hypothesized that not all RGB frames in a video are equally important and assigned a binary latent variable to each frame in training videos for indicating the importance  of a  frame within a  latent support vector machine model. Zaki \etal ~\cite{ZAKI2011572} proposed two new features with existing hand crafted features and developed the system using HMM based approach. Some have used appearance-based features and divided the approach into sub-units of RGB and tracking data, with a HMM model for recognition~\cite{Cooper:2012:SLR:2503308.2503313}. These methods either tackled few number of sign class variation with uniform background, trained models using a fraction of test subject's data or did not explore the hand shapes rigorously. 

Compared to RGB methods, skeletal data has received little attention in ASL recognition. However, in a closely related human action recognition task, a significant amount of work has been done using body joint information. Shahroudy \cite{7780484} released the largest dataset for human activity recognition. They proposed an extension of long short term memory (LSTM) model which leverages group motion of several body joints to recognize human activity from skeletal data. A different adaptation of the LSTM model was proposed by Liu \cite{8101019} where spatial interaction among joints was considered in addition to the temporal dynamics. 
Some researches focused on capturing salient motion pattern of body joints \cite{7410817} and some leveraged hierarchical properties of joint configuration \cite{7298714}.
Several attention based model were  proposed for human activity analysis \cite{song2016end,8226767}.
Few prior works converted skeleton sequences of body joints or RGB videos into an image representation and then applied state-of-art image recognition models to achieve good results \cite{DBLP:journals/corr/abs-1711-05941,newRepSkeleton}. In generic activities the whole body moves, which is not the case for ASL sign gestures where primarily the hands move. Estimated body joint poses (2D/3D) can be used to a certain level, because pose data only gives a high level motion pattern of a sign gesture. Hence, sign gestures recognition demands careful hand shape modeling.

Hand segmentation or recognition is also a well studied problem in computer vision \cite{egohands, oxford_hand, deephand}. Some of these methods concentrate on hand detection rather than modeling hand shapes \cite{oxford_hand}, some has different viewpoints such as egocentric views \cite{egohands}. Koller \etal \cite{deephand} trained a CNN for hand shape modeling in an semi supervised manner. This method is closer to a part of our work, however, it lacks fine grained hand shape modeling. On the other hand, our goal is to learn robust hand shape representation using careful supervision. 

We use only RGB modality in sign language classification. This makes our  system independent of depth sensor. 
However, unlike traditional RGB based methods which use feature based HMM, we base our model on deep learning methods considering the size and variation in the dataset. Besides, our method focuses on learning fine grained hand-shape features before the sign classification phase. It uses fine supervision from a small fraction of data to learn hand shape and further use sequential modeling using learned representation for final classification task. We show by leveraging careful supervision on hand shapes our methods can achieve significant performance boost.

\section{GMU-ASL51 Dataset}
All of our hand shape learning as well as ASL classification tasks were evaluated using GMU-ASL51 benchmark \cite{gmuASL}. We picked this dataset because it is the only publicly available dataset of this type (isolated word level ASL gestures) with large number of sign variation. GMU-ASL51 has 51 word level ASL signs performed by 12 subjects of different ages, gender and builds. The dataset was collected using depth sensor and has two modalities: RGB videos and 3D skeletal body parts. More detail can be found in the paper.  
In this dataset, only video level class label is available. One of our main contributions of this work is to systematically annotate per frame hand shape from each video.


\section{Our Approach}
We refer to our proposed sign gesture classification pipeline as FineHand. It has two parts. In this section, we start with describing the first part which is a CNN model trained to learn hand shape representation. Hereafter, we refer this model as hand shape model (or embedder). Then, we present the second part which is a sequential RNN classification model learn to recognize different sign gestures using hand shape embeddings.   
\subsection{Hand Shape Embeddings}
The goal of this part of our pipeline is to learn high-dimensional representation of hand shape which is discriminative for ASL gesture recognition. 
In this section we are first going to present how we crop hand patches using off the shelf pose estimation method, followed by iterative hand shape learning mechanism. Finally, we present some qualitative results from our learned hand shape model.
\label{pose_estimation}
\paragraph{Pose Estimation}
Pose estimation is the process of estimating 
2D or 3D body joint locations (e.g. wrist, elbow) in single image.
Typically it is done by first detecting human subjects in an image frame and then parsing body joint location\cite{1701.01779, Gkioxari:2014:UKD:2679600.2680056,1608.08526}.
or the inferring the body parts without first detecting the person as a whole \cite{Pishchulin_2016_CVPR, 1605.03170, cao2018openpose}. 
\begin{figure}[h]
	\centering
	\includegraphics[clip, trim= 0.5cm 20.0cm 7.5cm 1.0cm, width=\linewidth, height=.2\textheight]{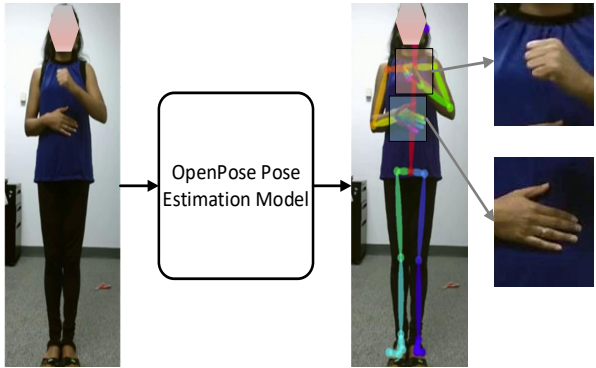}
	\caption{Pose estimation and hand cropping process.}
	\label{fig:hand_crop}
\end{figure}
For this work we have chosen the state-of-the-art 2D human body pose estimation approach OpenPose~\cite{cao2018openpose}. It is to be noted that we only use hand poses to crop a hand patch from each image. 
Figure \ref{fig:hand_crop} shows the whole process of estimating body poses from an RGB frame and cropping hand patches.


\begin{table}[h]
    \caption{Iterative hand-shape learning process. In the header row P, C and T symbolizes prediction, correct and total count respectively. Iter 1 is the manual annotation, hence all labels are correct. T column of final pass denotes the cumulative count of hand-shape samples for the class represented by rows. Iteration is abbreviated as Iter.}
	\centering
	\begin{tabular}{|c|c|c|c|c|c|c|c|}
		\cline{1-8}
		\multirow[]{2}{*}{Class}& \multirow[]{2}{*}{Iter 1} &\multicolumn{3}{c|}{\multirow[t]{2}{*}{Iter 2}} & \multicolumn{3}{c|}{\multirow[t]{2}{*}{Iter 3}} \\
		\cline{3-8}
		&  & P & C & T & P & C & T \\
		\cline{1-8}
		C1 & 402 & 598 & 534 & 936 & 524&511&1447 \\
		\cline{1-8}
		C2 & 217 & 277 & 277 & 494 & 281&277&771 \\
		\cline{1-8}
		C3 & 69 & 88 & 73 & 142 & 102&96&238 \\
		\cline{1-8}
		C4 & 328 & 554 & 408 & 735 & 435&396&1131 \\
		\cline{1-8}
		C5 & 163 & 236 & 196 & 359 & 219&198&557 \\
		\cline{1-8}
	\end{tabular}
	\label{tab:iter_table}
\end{table}

\subsubsection{Iterative Hand Shape Learning}
\label{it_hshape_learning}
GMU-ASL51 dataset has 12 subjects and 51 word level sign classes with only sign video level labels. 
In this phase, we learn per frame hand representation by training hand shape embedder. 
Our approach depends on some early manual hand shape annotation. To be more specific, 
We take one gesture sample per subject for each sign class which gives us a total of $612$ ($12 \times 51$) sign videos. We extract hand-shape patches using pose data and manually group them based on visual similarity into $41$ classes.
These examples are used to fine-tune ResNet-50 CNN architecture~\cite{resNet}.
In the second iteration, we use the high-confidence predictions of our initial model to predict hand shape patches from another 612 sign videos (different from the 612 set of first iteration).
At this point we will have some incorrect predictions because the model is trained on small amount of data. We manually correct the incorrect predictions. Although this fix is manual but we can see that significantly less amount of labor is needed in the second iteration than the first one. After this phase, we have more annotated hand-shape patches and we retrain our model. With additional 
iterations our model becomes more robust. 
\begin{figure}[h]
	\centering
	\includegraphics[width=.9\linewidth, height=.3\textheight]{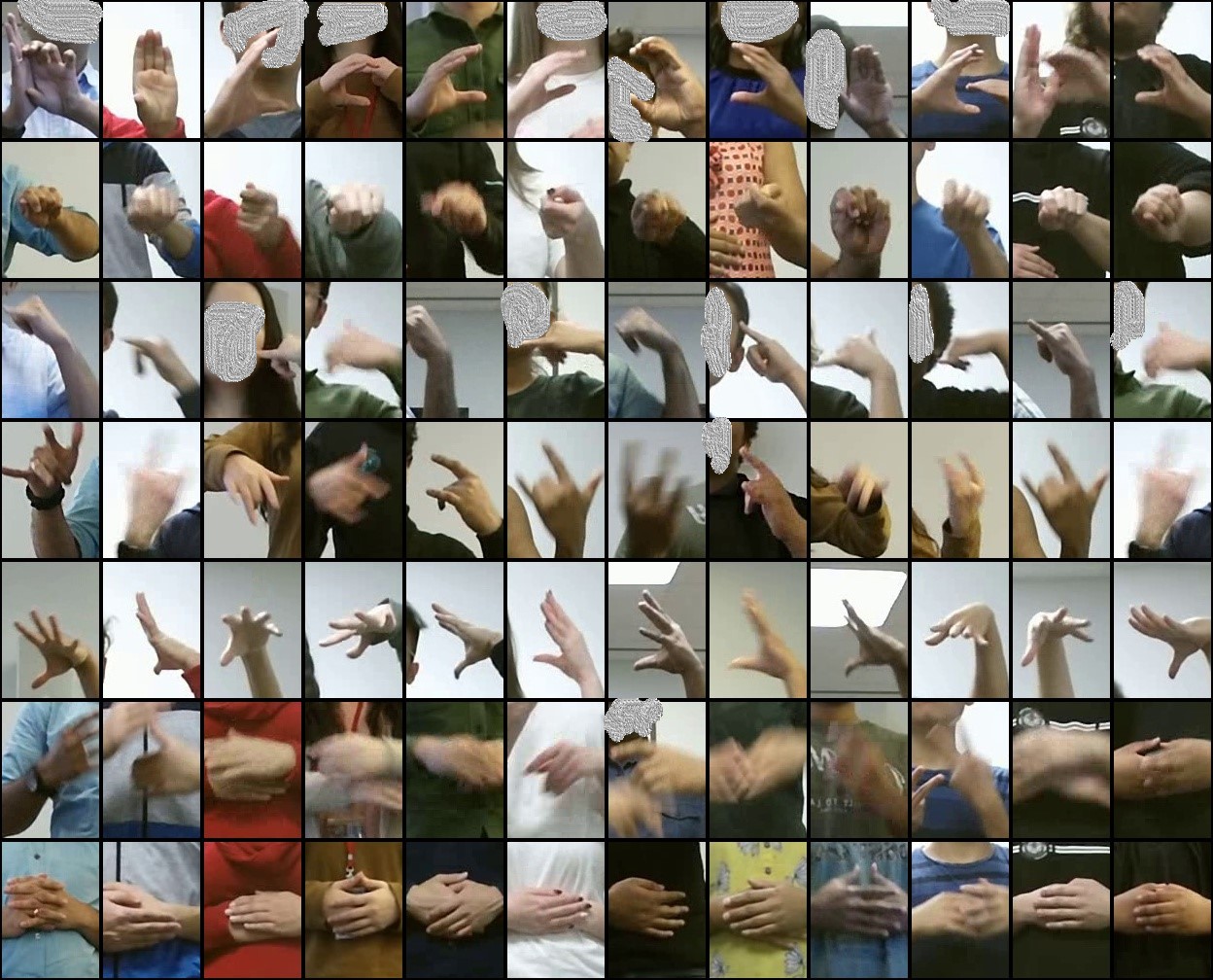}
	\caption{Sample hand-shapes. Each row shows $12$ randomly picked samples from the created dataset in the iterative hand shape learning phase.}
	\label{fig:hs_classes}
\end{figure}
Similarly we can start a third iteration and so on. 
We perform 3 such iterations and end up with $41$ classes of hand shapes which are distributed among $51$ sign gesture classes of GMU-ASL51 dataset. Table \ref{tab:iter_table} shows the count of annotated hand-shape samples for five classes and three iterations. Details for all $41$ classes will be provided in the supplemental materials.
It should be noted here, in three iterations the fraction of sign gesture data we use to train the model were $4.16\%$, $8.32\%$ and $12.5\%$ respectively. It should be also mentioned that, we use per frame hand-shape annotation on this fraction of data which means that a video gesture sample could possibly generate several hand-shape training examples. Effects of this incremental learning on sign classification is discussed in more detail in the result section.  
Among $41$ hand-shapes two are unusual: garbage and rest-position. Keeping those two classes is significant because, in a sign video most of the frames hands are blurred or are in a resting position. If we have a way to learn these uninformative hand shapes then we can exclude them during sign language modeling and hence have robust feature representation. 
Figure \ref{fig:hs_classes} shows examples of several hand shape classes picked from $41$ class hand shape dataset. Each row represents one class. Twelve samples in a row are picked randomly from our created hand shape dataset. We observe significant intra-class variation. Bottom two rows show the sample hand shapes from the class \texttt{garbage} and \texttt{rest-position} respectively. This ResNet based hand shape model is a module of the proposed FineHand sign gesture classification pipeline. After being trained on hand shape data, parameters of this model can be freezed during sequential learning of signs.
\begin{figure}[h]
	\centering
	\includegraphics[clip, trim = 5.5cm  9.5cm 6.5cm 7.5cm, width=\linewidth, height=.4\textheight]{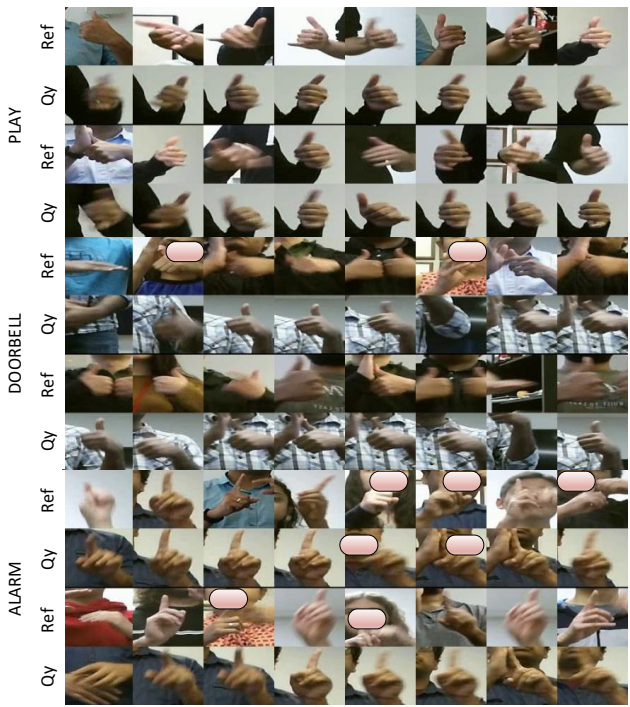}
	\caption{Predicted hand shape classes from a trained ResNet-50. For each of three sign classes two samples are shown where $Qy$ means query hand shape sequence and  $Ref$ means a reference sample of predicted label for each corresponding query hand patch from the training samples.}
	\label{fig:query_hs}
\end{figure}
\paragraph{Qualitative Results}
The hand shape model network is trained in an incremental fashion. Manual annotation in the first iteration and all the other generated hand shape labels are hand independent. 
This means if we look at all the samples from a class, we will find examples from both hands. Of course, a shape will be horizontally flipped or rotated as we look at the left versus right hand. First, second and fourth rows in Figure \ref{fig:hs_classes} show such examples.
Figure \ref{fig:query_hs} presents predicted and reference examples patches using a trained hand shape model, four top rows show two samples (divided by thin black line) from the sign \texttt{play}. 
\begin{figure}[h]
	\centering
	\includegraphics[clip, trim= 3.5cm 7.5cm 3.0cm 8.0cm, width=\linewidth, height=.3\textheight]{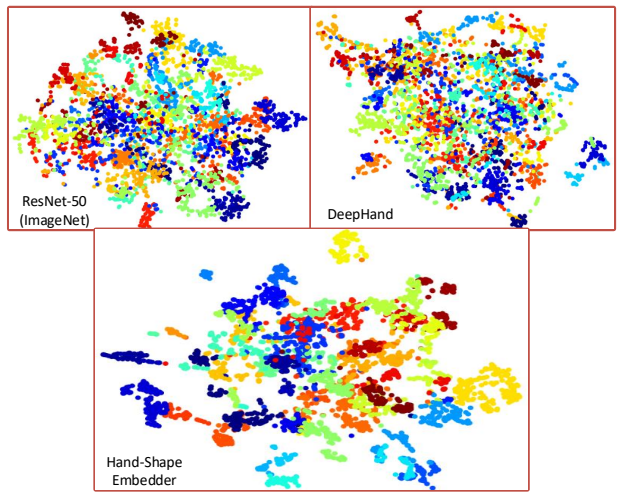}
	\caption{T-SNE visualization of hand shape embeddings from different model sources for 4033 samples. Top figures show representation obtained from ResNet-50 (ImageNet pre-trained) and DeepHand respectively, from left to right. Bottom figure shows embedding obtained from a trained (cross subject) ResNet model using hand shape labels created by us.}
	\label{fig:tsne_fig}
\end{figure}
For each samples the second row shows the query (Qy) hand patch and the corresponding patch in the row above (Ref) represents a reference training sample from predicted class. This classification model is trained on cross-subject manner which means none of the hand patches from the test subject (query) is used during training procedure. We observe, in most of the cases, hand shape model learns useful feature representation which is invariant to rotation, scaling and background. We will then keep the penultimate layer of the model to be high-dimensional embedding each hand patch. 

Figure \ref{fig:tsne_fig} shows the comparison among T-SNE representation of embeddings obtained by our model. It shows that hand shape representation learned by our CNN embedder cluster better than embeddings produced by ResNet (ImageNet trained)~\cite{resNet} or DeepHand~\cite{deephand} models.  This gives us the motivation behind using this effective representation in sign video classification task.
\begin{figure*}[h]
	\begin{center}
		\includegraphics[clip, trim= 4.5cm 10.0cm 4.5cm 8.5cm, width=.7\linewidth,height=.35\textheight]{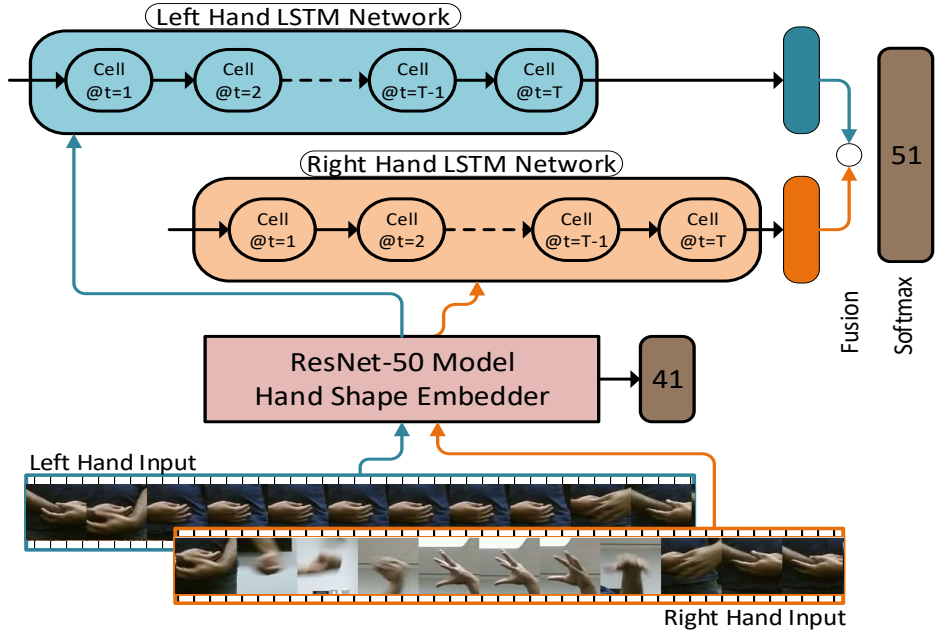}
	\end{center}
	\caption{FineHand RNN model. The ResNet-50 model is trained separately first on $41$ hand-shape classes. After training, it provides representation for each hand-patch video which is then used in sequential LSTM classifier for $51$ sign classes in the dataset.}
	\label{fig:arch}
\end{figure*}

\subsection{Sequential Sign Gesture Learning}
As pointed out and shown in the previous section, learned hand shape representation could be useful in classifying ASL gesture videos. However, sequential dynamics in video data still needs to be modeled carefully for better classification accuracy. With a trained hand shape embedder, each video can be converted into a sequence of embeddings. We base our sequential modeling using recurrent neural network (RNN). 

\paragraph{Recurrent Neural Network}
For modeling sequential data, recurrent neural network (RNN) 
have yielded impressive results on a variety of sequence prediction tasks~\cite{DBLP:journals/corr/Lipton15}. RNN models can capture temporal dynamics by maintaining an internal state. However, the basic RNN has problems dealing with long term dependencies in data due to the vanishing gradient problem. 
Some solutions to the vanishing gradient problem involve 
careful initialization of network parameters or  early stopping  \cite{DBLP:journals/corr/abs-1211-5063}. But the most effective solution is 
to modify the RNN architecture in such a way that it has a memory state (cell state) at every time step that can identify what to remember and what to forget. This architecture is referred as
long short term memory (LSTM) network.  
While the basic RNN is a 
direct transformation of the previous state and the current input, the LSTM maintains an internal memory and  has a 
mechanism to update and use that memory. This is achieved by deploying four separate neural networks also 
called gates. More detailed description of LSTM model can be found in~\cite{Hochreiter:1997:LSM:1246443.1246450}.

\subsubsection{FineHand Architecture}
We propose to use an LSTM recurrent neural network for this task. For a sign video, input to this model is a sequence of embeddings  obtained from our ResNet based hand shape embedder model. 
Assume we have a sequence cropped hands images $R^{F \times H \times W}$ where $F, H, W$ are number of frames, height and width of cropped hand patches respectively. Feeding the images to the hand shape model we obtain $D=2048$ dimensional embedding for each frame and output of the form $R^{F \times D}$. 
We use this data to train an LSTM model where $F$ is the number of temporal steps. For simplicity to deal with different number of frames, we sample $T$ predetermined number of frames uniformly. Hence, input to the LSTM network is $R^{T \times D}$. 
Finally we pick the hidden state at the end of last layer of LSTM network and take that as a encoded representation for each sequence. This final representation captures rich temporal as well as spatial hand shape features and is fed to a fully connected neural network layer to produce prediction probability distribution for sign gesture video classification. 
We train two different networks for left and right hands. We fuse the output of two networks at the end to produce final classification scores. Figure \ref{fig:arch} shows the details of our proposed architecture. It shows that the left-hand and right-hand patches are input to the trained hand-shape embedder model and the generated representation is being used as input for the recurrent LSTM networks. 

\begin{table*}
	\begin{center}
		\caption{Test accuracies across 12 subjects. In header row each subject is represented by S appended with subject number. Bottom row shows the result of our proposed FineHand architecture. Other rows are different comparative methods. 3D labeled three rows show the results using DeepHand embedding with Kinect 3D pose. 2D labeled rows show the similar experiments with OpenPose poses.}
		\begin{tabular}{|c|c|ccccccccccccc|}
			\cline{3-15}
			\multicolumn{1}{c}{}&\multicolumn{1}{c|}{}&S01 & S02 & S03 &S04 & S05 & S06 & S07 &S08 & S09 & S10 & S11 &S12 & Average\\
			
			\cline{2-15}
			\multicolumn{1}{c|}{}& 3D CNN &0.62	&0.63	&0.60	&0.60	&0.56	&0.56	&0.62	&0.45	&0.39	&0.51	&0.46	&0.17	&0.51 \\		
			\multicolumn{1}{c|}{}& FoaNet\cite{foanet} Style &0.18 &0.22 &0.10 &0.19 &0.11 &0.28 &0.15 &0.17 &0.09 &0.12 &0.11 &0.05 &0.15 \\
			\hline
			\multirow{3}{*}{\rotatebox{90}{3D}} 
			&PoseLSTM &0.81	&0.88	&0.68	&0.84	&0.88	&0.85	&0.85	&0.78	&0.81	&0.83	&0.76	&\textbf{0.83}	&0.82 \\
			&RgbLSTM &0.81	&0.82	&0.85	&0.89	&0.81	&0.89	&0.93	&0.69	&0.83	&0.72	&0.81	&0.37	&0.78 \\
			&FusionLSTM &0.90	&0.93	&0.88	&0.93	&0.90	&\textbf{0.95}	&0.96	&0.84	&0.91	&0.89	&0.89	&0.67	&0.88 \\
			\hline
			\multirow{3}{*}{\rotatebox{90}{2D}}
			&PoseLSTM &0.83	&0.90	&0.89	&0.92	&0.90	&0.92	&0.95	&0.88	&\textbf{0.94}	&0.93	&0.94	&0.63	&0.89 \\
			&RgbLSTM &0.80	&0.82	&0.88	&0.93	&0.82	&0.92	&0.95	&0.79	&0.90	&0.75	&0.92	&0.27	&0.81 \\
			&FusionLSTM &0.85	&0.85	&0.93	&\textbf{0.95}	&0.87	&0.92	&0.98	&0.83	&0.93	&0.85	&0.96	&0.40	&0.86 \\
			\hline
			\multicolumn{1}{c|}{}& FineHand (ours) &\textbf{0.93} &\textbf{0.98}	&\textbf{0.97} &0.91	&\textbf{0.91} &0.93	&\textbf{0.99} &\textbf{0.88}	&0.91 &\textbf{0.94}	&\textbf{0.96} &\textbf{0.83} &\textbf{0.93} \\
			\cline{2-15}
		\end{tabular}
		\label{tab:result_table}
	\end{center}
\end{table*}

\subsubsection{Training Details}

Average cross entropy loss is used to update the network parameters for a batch. Given a true one hot encoded class label of $y_{i,j}$ and corresponding predicted score of $\hat{y}_{i,j}$ this loss is calculated as Equation \ref{eq:loss}.
\begin{equation}
\label{eq:loss}
L = -\frac{1}{N}\sum_{i=1}^N\sum_{j=1}^C y_{i,j}\log{(\hat{y}_{i,j})}
\end{equation}
Here, $N$ is the size of the minibatch and $C$ is the number of classes. For our experiments, these values are $64$ and $51$ respectively.
For selecting the number of layer of this network, hidden state size and time steps of input, we performed grid search. We found best validation accuracy is obtained with 2 layer networks, 512 as hidden state size and 20 (T) as the input sequence length. Size of the input dimension at each time steps is the size ($2048$) of the representation produced by hand shape embedder.  We used Adam Optimizer for training our networks~\cite{DBLP:journals/corr/KingmaB14} with learning rate set to $0.0001$.

\section{Experiments}
In this section, we are going to describe different methods we used in comparison with our proposed architecture.
\subsection{Comparative Methods}
\label{compare_methods}
\paragraph{3D Convolution}
Convolutional Neural Network (CNN) with 3D convolutional kernel (3D CNN) has shown promising performance in classifying human activities in video \cite{Ji:2013:CNN:2412386.2412939}. That's why we chose 3D CNN on RGB hand patch videos for one of our baselines. It consists of four 3D convolutional layers and two fully connected layers at the end.
There are two separate networks for left and right hands' patches. Final embedding of these two networks are concatenated before producing softmax
score.


\paragraph{Deep Hand Model}
The authors in~\cite{deephand} trained a 22 layer deep convolutional neural network (CNN) with more than 1 million images, from videos of Danish and New Zealand sign language. The data is weakly labeled with only video level annotation. The CNN model for estimating the likelihood of hand shapes, is trained using EM algorithm, jointly with Hidden Markov Model (HMM) for parsing sign gestures.  
The network is trained to recognize 60 hand shape classes plus one garbage class which determines the start or end of a video. 
We forgo the softmax classification layer of this network and use the embeddings computed by this pre-trained model as representations of crops of hand patches from ASL data. The final layer embedding (1024 dimensional) as a feature vector. We use representation generated by this model in our comparative experiments described next.
\paragraph{Kinect 3D Pose}
For this experiment we used RGB video and 3D skeletal pose data as computed by Kinect sensor \cite{Zhang:2012:MKS:2225053.2225203}.
We do three types of experiments by using these two types of data. Two of those experiments use each modality separately. The other one uses fusion strategy to get maximum out of both modalities. These models are referred as 3D version of PoseLSTM, RgbLSTM and FusionLSTM in result section. We use embedding from Deep Hand model in this comparative method. 


\paragraph{OpenPose}
This experiment is similar to the process described in last paragraph except only uses estimated poses from RGB videos instead of using 3D skeletal poses generated by Kinect sensor. Process of estimating poses from video is described in section \ref{pose_estimation}. Rationale behind doing this experiment is to exclude the dependency on depth sensor. Since, this poses are estimated from RGB video, this experiment depends solely on RGB modality. These models are referred as 2D version of PoseLSTM, RgbLSTM and FusionLSTM in result section.
\paragraph{Comparison with Similar Work}
It was difficult to find a similar work which can be directly compared to our method. This is due to the nature of our set up which is isolated ASL word level sign recognition. We could not find any standard public dataset other than recently released GMU-ASL51 for this purpose. However, there is a public dataset on generic gestures \cite{IsoGd}. 
Various deep learning based methods have been proposed to model generic isolated gestures \cite{8265571,8265580,8265581,foanet}. Narayana \etal proposed FOANet which uses fusion of different channels on cropped hand patches and full body \cite{foanet}. Using different modalities (RGB, depth and flow) with those channels this work sparsely fuses 12 channels of inputs to model gestures. Details can be found in the paper.
We tried to reproduce this work on GMU-ASL51 as closely as possible. However exact recreation was not possible due to several factors such as number of data channels used, number of location features of hand patches and mechanism of cropping hand patches. Details of these differences will be provided in supplemental materials. 

\section{Results}
Table \ref{tab:result_table} shows our experimental results. All of our experiment shown are cross subject in manner. For a particular test subject, we trained our model using data from all other subjects in the dataset. 
 This cross subject evaluation criteria supports practical usability of our system to subjects unknown to the trained model. Each column in Table \ref{tab:result_table} shows test accuracy of one subject in GMU-ASL51. First two rows shows the baseline results: 3D CNN and FoaNet style implementation.  
Next two blocks of three rows show experiments with 3D and 2D poses respectively. Finally, bottom row shows results of our proposed method (FineHand).

Result shows that methods using 2D poses achieve almost similar performance to 3D Kinect poses ($88\%$ vs $86\%$) which is interesting because 2D pose methods depend only on RGB data. It should be noted that, unlike Kinect sensor, OpenPose provides finger joints. We presume, even though these poses lack depth information, finger joints help to achieve comparable performance with 3D Kinect poses.

From Table \ref{tab:result_table} we observe that, our proposed method, FineHand outperforms top models using 3D and 2D poses by $5\%$ and $7\%$ respectively. It should be mentioned that, pose based models use pose data while FineHand model only uses RGB hand patches. Taking this into consideration, it is fair to compare FineHand  with RGB only versions (RgbLSTM) of 3D and 2D implementation. In that case, FineHand outperforms those implementations by $15\%$ and $11\%$ respectively. This significant boost in classification accuracy can be justified by the way FineHand learns hand shapes. Representation used for RgbLSTMs is taken from DeepHand, a pre-trained model on huge amount of hand shapes data from a different class distribution as des \ref{compare_methods}. On the other hand, FineHand embedder learns representation from of fine grained hand shapes which has proven to be crucial for this kind of significant performance gain in classification. Our best method outperforms the work came with the dataset \cite{gmuASL} by $12\%$.

The FOANet style implementation on GMU-ASL51 has really bad performance (second row in Table \ref{tab:result_table}) even though it is one of the top performing models for generic gestures. One possible reason is the number of data channels used. While original work uses $12$ channels, in our implementation we use only $2$ channels to make it comparable with our proposed work. Another reason is the training procedure. FOANet architecture proposed to capture sequential dynamics in a video gesture by using sliding window based approach where classification scores for a video gesture was computed by taking averages over all sliding window scores. Our method however, pre samples some fixed number of frames from a video and produces one set of prediction score per video gesture.
\begin{table}[h]
	\begin{center}
	    \caption{Average cross subject sign recognition accuracy on different iterations of hand shape learning}
		\begin{tabular}{|l|c|}
			\hline
			Iterations (\% Train Data) & Accuracy \\
			\hline\hline
			Iteration 0 (0.00\%) & 0.65 \\
			Iteration 1 (4.17\%) & 0.89 \\
			Iteration 2 (8.32\%) & 0.91 \\
			Iteration 3 (12.5\%) & 0.93\\
			\hline
		\end{tabular}
	\end{center}
	\label{tab:it_table}
\end{table}
\paragraph{Effect of Hand-shape Learning Iterations}
In section \ref{it_hshape_learning}, we briefly described how hand-shape learning CNN is trained in successive iterations. 
We hypothesize, increasing these iterations will boost up the sign classification accuracy. Table \ref{tab:it_table} shows the average recognition accuracy on using different fractions of data for training the embedder CNN. Here, `Iteration 0' represents no hand-shape learning meaning we use embedding representation from ImangeNet pre-trained CNN model as input to LSTM network for sign classification.

\begin{table}[h]
	\begin{center}
	    \caption{Average cross subject sign recognition accuracy for different scenarios of hand usage.}
		\begin{tabular}{|l|c|}
			\hline
			Input Types & Accuracy \\
			\hline\hline
			Left Hand & 0.65 \\
			Right Hand & 0.90 \\
			Both Hands (max score) & 0.86 \\
			Both Hands (catenation) & 0.92 \\
			Both Hands (mean score) & 0.93 \\
			\hline
		\end{tabular}
	\end{center}
	\label{tab:hands_table}
\end{table}
\paragraph{Both Hands vs Single Hand}
We are also interested to compare results obtaining from either using left or right hand and using them together. Usually some signs are dominated by single hand while others are double handed. 
It is obvious that, using both hands' information will increase the accuracy. However, we want to see how much improvement is possible using both hands. In case of using both hands, we also show if there is any best fusion mechanism. Table \ref{tab:hands_table} shows this results. We observe that, good accuracy is achieved using only right hand input. 
This is because, we notice that, all of the subjects use right hand as dominant hand in GMU-ASL51 dataset. However, using both hands we have $3\%$ improved accuracy which suggests that, in some cases right hand is also important. Among the fusion strategies we found, averaging logits works best.
 \begin{table}[h]
 	\begin{center}
 	\caption{Average cross subject sign recognition accuracy for different learning mechanisms (joint vs separate).}
 		\begin{tabular}{|l|c|}
 			\hline
 			Train Type & Accuracy \\
 			\hline\hline
 			Separate Learning & 0.93 \\
 			Joint Learning & 0.89 \\
 			\hline
 		\end{tabular}
 	\end{center}
 	\label{tab:learning_table}
 \end{table}
\paragraph{Effect of Joint Learning}
This section shows comparison between learning the whole network jointly vs separately. Our default set up is separate learning where first we train the hand shape CNN model using annotated hand patch data, then freeze it during sequential learning of embedding produced by it on training video hand patches. In case of joint learning, we don't freeze the hand shape CNN network during sequential sign learning. We notice that, joint learning worsen the performance of the whole network. We presume, since the CNN is first trained on hand-shape supervision, it might get confused when sign level gradient updates are done on its parameters. Also, during back propagation gradient has to travel backwards through the LSTM network before hitting hand-shape CNN model. This causes derogatory updates on CNN parameters. Hence, produced embedding representation differs in each iteration, which impact LSTM learning negatively.

\section{Conclusion and Future Work}
We have demonstrated effectiveness of learning hand-shape representation for ASL sign gesture classification from video. We showed qualitative results of better representation produced by our proposed hand-shape learning mechanism. We also verified that, this representation can achieve superior sign classification accuracy than other sources of embedding learning. Our proposed method is RGB only but outperforms multi-modal (RGB and pose) approaches of sign language recognition. Given the hand shape annotation, exploring sentence level sign language modeling is an interesting direction. We believe, per frame hand shape annotation will help to localize individual sign gestures in a sign video sentence.

{\small
\bibliographystyle{ieee}
\bibliography{submission}
}

\end{document}